\documentclass[times, review, 10pt]{elsarticle}
\usepackage{geometry}
\usepackage{amssymb}
\usepackage{subfigure}
\usepackage{longtable}
\usepackage{multirow}
\usepackage{mathrsfs}
\usepackage{amsfonts}
\usepackage{amsmath}
\usepackage{array}
\usepackage{color}
\usepackage[ruled]{algorithm2e}
\usepackage{float}
\biboptions{numbers,sort&compress}

\journal{Pattern Recognition}

\begin{document}

\begin{frontmatter}



\title{Precise Facial Landmark Detection by Dynamic Semantic Aggregation Transformer}


\author[firstauthor,firstauthor2]{Jun Wan}
\author[firstauthor]{He Liu}
\author[secondauthor]{Yujia Wu\corref{cor1}}
\author[firstauthor2]{Zhihui~Lai}
\author[fourthauthor]{Wenwen Min}
\author[thirdauthor]{ Jun~Liu}

\cortext[cor1]{Corresponding author: Yujia Wu (wuyujia@whu.edu.cn).}
\address[firstauthor]{School of Information Engineering, Zhongnan University of Economics and Law, Wuhan, 430073, China}
\address[firstauthor2]{College of Computer Science and Software Engineering, Shenzhen University,  Shenzhen, 518060, China}
\address[secondauthor]{ School of Information Science and Technology, Sanda University, Shanghai 201209, China}
\address[fourthauthor]{School of Information Science and Engineering, Yunnan University, Kunming, Yunnan 650091, China}
\address[thirdauthor]{Information Systems Technology and Design Pillar, Singapore University of Technology and Design, Singapore, 487372, Singapore}

\begin{abstract}

At present, deep neural network methods have played a dominant role in face alignment field. However, they generally use predefined network structures to predict landmarks, which tends to learn general features and leads to mediocre performance, e.g., they perform well on neutral samples but struggle with faces exhibiting large poses or occlusions. Moreover, they cannot effectively deal with semantic gaps and ambiguities among features at different scales, which may hinder them from learning efficient features. To address the above issues, in this paper, we propose a Dynamic Semantic-Aggregation Transformer (DSAT) for more discriminative and representative feature (i.e., specialized feature) learning. Specifically, a Dynamic Semantic-Aware (DSA) model is first proposed to partition samples into subsets and activate the specific pathways for them by estimating the semantic correlations of feature channels, making it possible to learn specialized features from each subset. Then, a novel Dynamic Semantic Specialization (DSS) model is designed to mine the homogeneous information from features at different scales for eliminating the semantic gap and ambiguities and enhancing the representation ability. Finally, by integrating the DSA model and DSS model into our proposed DSAT in both dynamic architecture and dynamic parameter manners, more specialized features can be learned for achieving more precise face alignment. It is interesting to show that harder samples can be handled by activating more feature channels. Extensive experiments on popular face alignment datasets demonstrate that our proposed DSAT outperforms state-of-the-art models in the literature. {\color{red}}Our code is available at  https://github.com/GERMINO-LiuHe/DSAT.

\end{abstract}


\end{frontmatter}


\section{Introduction}
Facial landmark detection, or known as face alignment, refers to detecting a set of predefined landmarks on 2D human face images. Its broad spectrum of usage across various domains has garnered significant interest in the academic community, including face recognition \cite{Wright2009RobustFR, He2024EnhancingFR}, face synthesis \cite{Yang2022HeterogeneousFR} and expression recognition \cite{liu2023cross, Savchenko2022ClassifyingEA}.

\begin{figure}[!t]
	\begin{center}
		\includegraphics[width=0.8\linewidth]{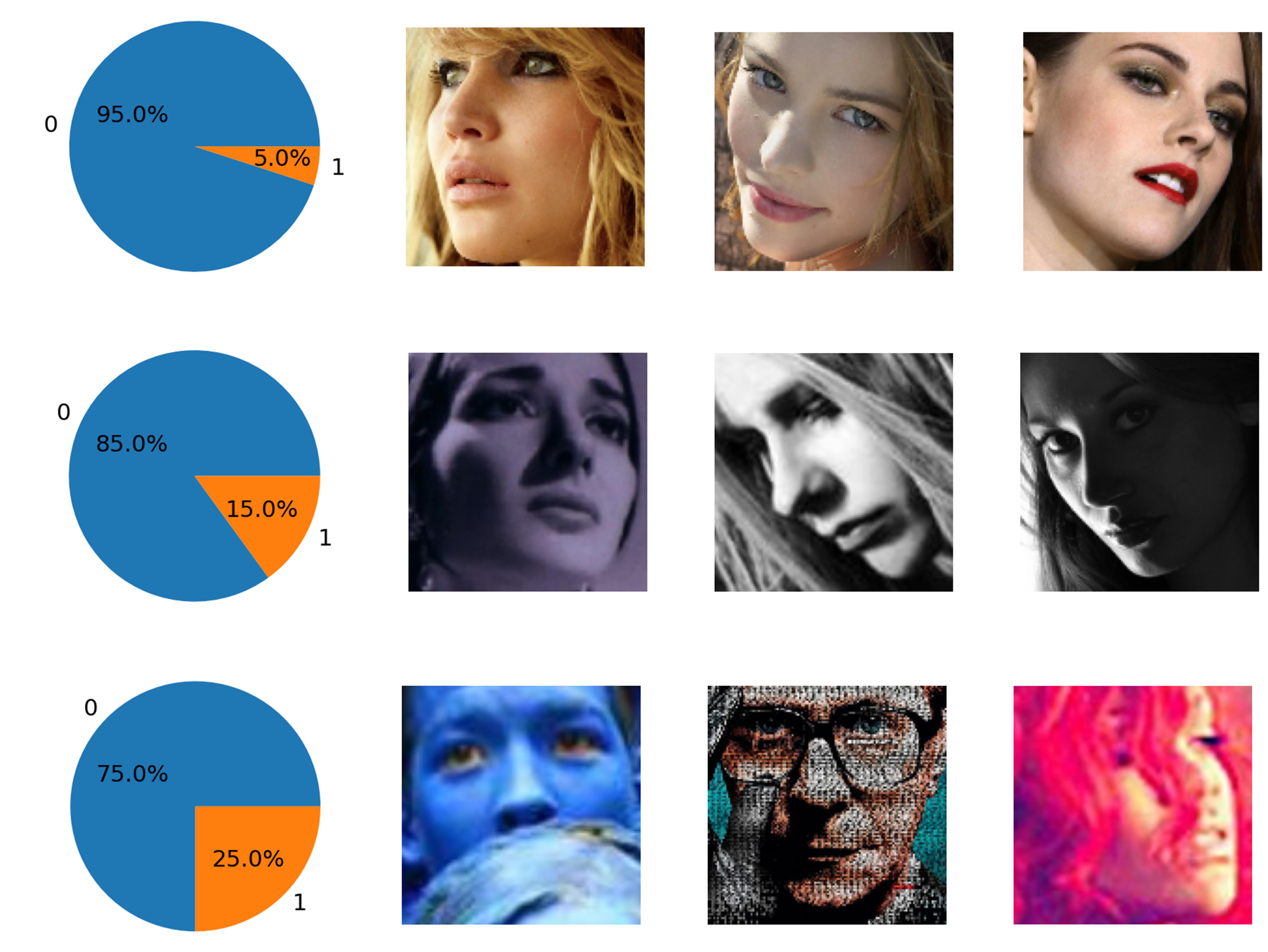}
	\end{center}
	\caption{The ratios of activated channels for different samples. The harder the face image, the greater the ratio of activated channels. This indicates that our proposed DSAT could learn specialized features according to the difficulty estimation of the face images for achieving more precise face alignment. }
	\label{problem}
\end{figure}

With the continuous development of convolutional neural networks (CNNs), facial landmark detection has achieved promising results \cite{Zhao2019MobileFANTD, wu2024diffusion, tian2024adaptive}. However, in practical applications, they need to face more complex environments, and their stability and accuracy are still greatly challenged. For example, faces with large poses or occlusions will significantly increase the difficulty of landmark detection, since the shape of the face will be very different from a neutral face. However, in most existing landmark detectors, all samples are trained by a predefined model architecture and share the same computational path. This makes the model parameters tend to be optimized with data from easy samples rather than hard samples  \cite{Johnson2019SurveyOD}, leading to mediocre feature learning and low detection accuracy. Moreover, current landmark detectors can not effectively deal with semantic gaps and ambiguities among features at different scales, which deteriorates the representative probability and reduces the alignment accuracy. Inspired by the fact that Dynamic Neural Networks (DNN) \cite{ Han2021DynamicNN, Xie2019OnGR} learn more efficient representations and have achieved remarkable results, this work explores how to improve face alignment performance by introducing Dynamic Neural Networks (DNN).

\begin{figure*}[!t]
	\begin{center}
		\includegraphics[width=0.92\linewidth]{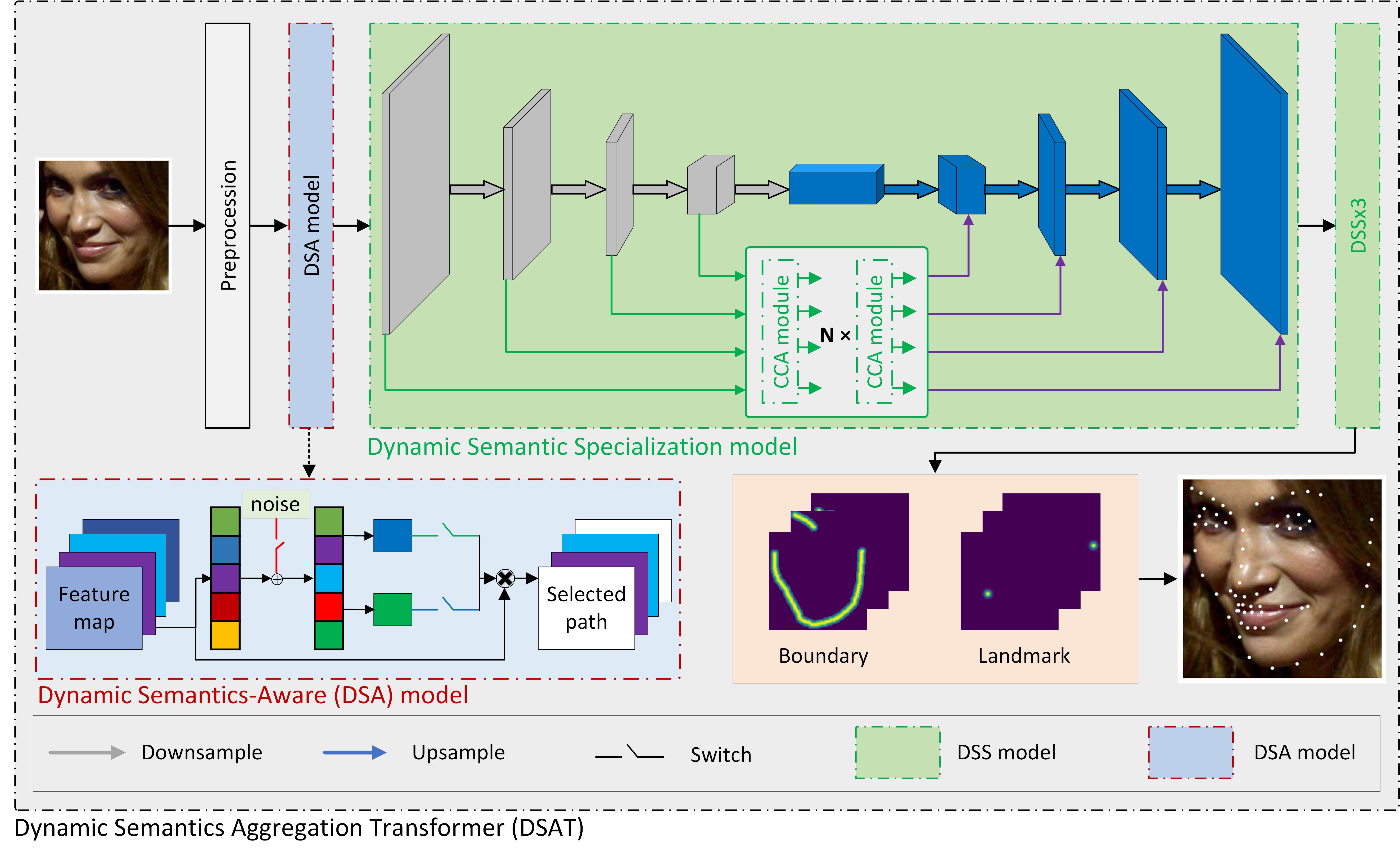}
	\end{center}
	\caption{The overall architecture of the proposed Dynamic Semantic Aggregation Transformer (DSAT). The proposed DSAT embeds the Dynamic Semantic-Aware model and Dynamic Semantic Specialization model in dynamic architecture and parameter manner to learn more effective specialized features for achieving more precise face alignment.}
	\label{overview}
\end{figure*}

We propose a Dynamic Semantic-Aggregation Transformer (DSAT) which embeds ideas of both dynamic architecture and dynamic parameters from DNN for more specialized feature learning. The key idea of DSAT is that we first partition the samples into subsets with similar faces, and then learn features from each subset. As the learned features are more focused on similar samples (as shown in Fig. \ref{problem}), they are more discriminative and expressive and we call them \textbf{specialized features}. To be specific, a novel Dynamic Semantic-Aware (DSA) model is first designed to partition all samples into subsets and activate specific pathways for them by estimating the semantic correlations of feature channels. Then, a Dynamic Semantic Specialization (DSS) model is proposed to mine the homogeneous information from features at different scales by querying them while alleviating their semantic gaps and ambiguities. Finally, by integrating the DSA model and DSS model into a novel Dynamic Semantic-Aggregation Transformer (DSAT) (as shown in Fig. \ref{overview}) through both dynamic architecture and dynamic parameter ways, our proposed DSAT can learn more effective specialized features for achieving more precise face alignment. The main contributions of this paper are summarized as follows: 

1) A novel Dynamic Semantic-Aware (DSA) model is designed to partition sample subsets and activate the specific pathways for them by estimating the semantic correlations of feature channels, which helps learn more specialized features for enhancing the representation ability and improving face alignment accuracy.

2) With a well-designed Cross-Channel Attention (CCA) module, a Dynamic Semantic Specific (DSS) model is proposed to mine homogeneous information between them by querying features at different scales, thereby making up the semantic gap between them and alleviating the semantic ambiguity.

3) We integrate the above two models into a Dynamic Semantic-Aggregation Transformer through both dynamic architecture and dynamic parameter ways to handle face alignment in the wild. With more effective specialized representations, our proposed DSAT outperforms state-of-the-art methods on challenging benchmark datasets such as COFW, 300W, AFLW and WFLW.

The rest of this paper is organized as follows. In Section 2, we elaborate on the summary of related work. In Section 3, we describe the details of our proposed Dynamic Semantics Aggregation Transformer model including the Dynamic Semantic-Aware model and Dynamic Semantics Specialization model. In Section 4, we evaluate the performance of the proposed method and compare it with state-of-the-art methods on benchmark datasets. Finally, in Section 5 there is a conclusion about this paper.

\section{Related Work}

The traditional face alignment methods mainly relied on model-based methods \cite{Cootes1995ActiveSM, Cootes1998ActiveAM}, which uses classical machine learning algorithms to directly predict facial landmarks from local features. With the advancements in deep learning, a new generation of face alignment methods has emerged, which can be generally categorized into two major groups: coordinate regression methods \cite{ZHU2022108325, liang2024generalizable} and heatmap regression methods \cite{Zhou2023STARLR, wan2023precise}.

\subsection{Model-based methods	}
Face alignment research has its origins dating back to the 1990s. Parametric models are utilized to constrain shape variations to improve face alignment performance. Cootes et al. propose the Active Shape Model (ASM) \cite{Cootes1995ActiveSM}, a statistical model designed for aligning facial landmarks. It includes shape and texture models, which iteratively refine the landmark positions to achieve alignment with a target configuration. Active Appearance Model (AAM) \cite{Cootes1998ActiveAM} utilizes principal component analysis (PCA) to establish a global facial shape model and a facial appearance model. In the detecting procedure, AAM is used for precise estimation of facial landmark positions by aligning the acquired appearance and shape model with the test image in a fitting process. In addition, many other model-based face alignment methods \cite{Cristinacce2006FeatureDA, Liu2009DiscriminativeFA} have been developed. However, the model-based methods have limited accuracy for faces with large pose or expression variations.

\subsection{Coordinate regression methods}
Coordinate regression methods directly predict the facial landmarks coordinates from input images using Convolutional Neural Networks (CNNs) or other deep learning methods. Lin et al. \cite{Lin2021StructureCoherentDF} adopt a graph convolutional network (GCN) to enforce the structural relationships among landmarks that can eliminate the noise between landmarks. Zhu et al. \cite{ZHU2022108325} propose a structural relation network (SRN) for occlusion-robust landmark localization, which aims to capture the structural relations among different facial parts. Xu et al. propose the AnchorFace \cite{Xu2020AnchorFaceAA} to handle faces with large poses, which aims to learn a set of anchor faces that can represent a wide range of facial variations and uses these anchor faces as references to regress landmarks. Recently, Xia et al. present a new framework \cite{Xia2022SparseLP} to address face alignment by utilizing a convolutional neural network (CNN) backbone combined with a transformer. This framework adopts a coarse-to-fine approach, allowing the model to learn the intrinsic relationships among facial landmarks and consequently improving the accuracy of the alignment results. Peng et al. \cite{Yang2024LDDMMFaceLD} introduce new angle features to describe the angle of different organs such as the chin and eyes, which help boost the face alignment accuracy. Liang et al. \cite{liang2024generalizable} propose to use the warping information to learn a generalizable face landmarker, making it applicable to facial images with different styles. Coordinate regression methods usually require a large number of samples for training. In addition, they are susceptible to interference and still suffer from hard faces.

\subsection{Heatmap regression methods.}
The heatmap regression method predicts landmark coordinates by generating landmark heatmaps and assuming that the location with the maximum value in the heatmap corresponds to the location of the landmark. With the widespread of CNNs, they have achieved promising performance. For example, Adaptive Wing loss \cite{Wang2019AdaptiveWL} is a new loss function for the heatmap regression method, which emphasizes paying more attention to small errors around the foreground pixels but reducing the impact of background pixels. Ma et al. present DSCN \cite{Ma2022RobustFA}, which introduces capsule networks to address facial landmark detection problems under occlusion and large pose. The ADNet \cite{Huang2021ADNetLE} is a novel face alignment approach that leverages an error-bias loss function that penalizes errors in the normal direction of the facial landmarks, leading to more accurate results. Wan et al. propose a novel Reference
Heatmap Transformer \cite{wan2023precise} to improve facial landmark detection accuracy by introducing reference heatmap information. Zhou et al propose a STAR loss \cite{Zhou2023STARLR} to alleviate the semantic ambiguity problem and thereby improving the alignment accuracy.

So far, current heatmap regression face alignment models can accurately locate the landmarks for neutral faces, but when encountering faces with large poses or severe occlusions, the alignment accuracy is still far from meeting the actual needs as 1) they use all samples to train and update the network parameters, which makes them tend to deal with general samples rather than harder ones, and 2) they usually use multi-scale features to improve prediction accuracy but they cannot effectively handle the differences and inherent relation between features at different scales. Our proposed DSAT can address the above issues by partitioning all samples into subsets and then learning specialized features from each subset to enhance the representation ability. Moreover, the Cross-Channel Attention (CCA) module in the DSS model can help to make up the semantic gap and alleviate semantic ambiguity. Therefore, our proposed DSAT achieves more precise face alignment.
\section{Network}
In this section, we will introduce the motivation and formulation of the proposed Dynamic Semantic Aggregation Transformer, which includes Dynamic Semantic-Aware (DSA) model and Dynamic Semantic Specialization (DSS) model. 

\begin{figure*}[!t]
	\begin{center}
		\includegraphics[width=0.96\linewidth]{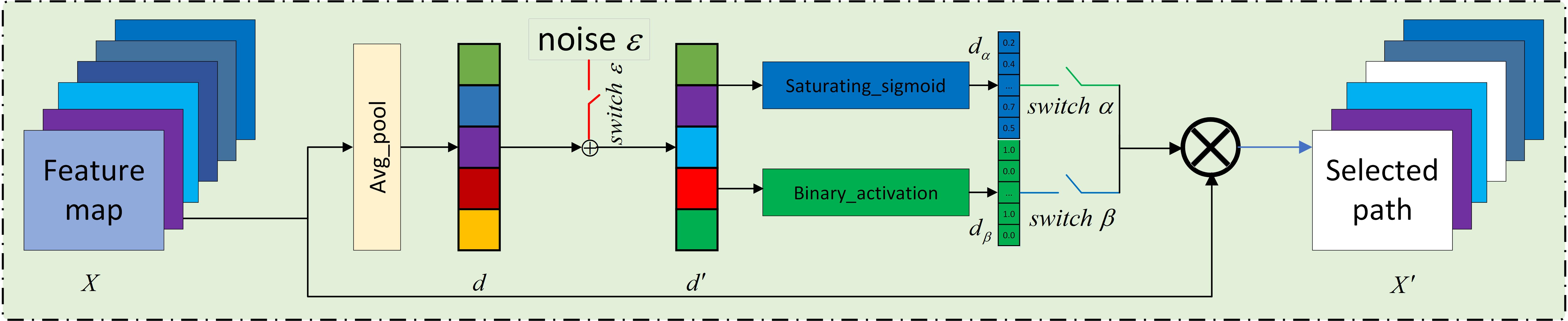}
	\end{center}
	\caption{The architecture of the proposed Dynamic Semantic-Aware (DSA) model. The DSA model is able to partition all training samples into subsets and activate specific pathways for each sample subset by estimating the semantic correlations of feature channels, which helps to learn more specialized features for enhancing representation ability.}
	\label{DSAM}
\end{figure*}

\subsection{Dynamic Semantic-Aware model}
Current deep-learning architectures have shown great success in face alignment. However, the calculation process of the model is predefined, i.e., all input samples share the same computational path. This makes face alignment models tend to learn more general features, which lacks flexibility and reduces alignment accuracy. In practice, different samples contain semantic content of different complexity, and samples with sufficient similarity (i.e., sample subsets) can share the same calculation path, the calculation paths of different sample subsets should be different. Hence, dividing all training samples into subsets encourages face alignment models to learn more specialized features and improves the alignment accuracy. Inspired by that dynamic neural networks can dynamically adjust their architecture according to the estimation of samples, we propose a Dynamic Semantics-Aware (DSA) model (as shown in Fig. \ref{DSAM}) to partition all training samples by estimating the semantic correlation between channels of sample features. DSA model activates specific pathways for each sample subset, which helps to learn more specialized features for enhancing representation ability and improving the face alignment accuracy.

The face image feature $X$ can be obtained by inputting face images into a preprocessing module as shown in Fig. \ref{overview}, and the preprocessing module contains serval convolution, max pooling and batch norm operations. We employ $W$, $H$ and $C$ to represent the width, height and channels of the feature maps respectively, and $X \in {R^{W \times H \times C}}$. Then, an adaptive-average-pooling operation is used to obtain a vector $d$, and the whole process can be formulated as follows:
\begin{equation}
	d = adapt\_avg\_pool(X)
\end{equation}where $d \in {R^{1 \times c}}$, and $adapt\_avg\_pool$ means the adaptive-average-pooling operation. After that, $d$ is used to estimate the semantic relevance between channels of sample features for sample division, i.e., the DSA model dynamically activates channels for each sample according to their semantic relevance. The weight of activated channels will be set to 1 and the other to 0, which can be regarded as the binary gates (denoted by $d''$). After obtaining the binary gates, the final output $X'$ of the proposed DSA model can be expressed as follows:
\begin{equation}
	X' = X*d''
\end{equation}

If samples have similar activation channels (i.e. binary gates), they can be considered as belonging to the same subset of samples. They share the same computational path, so they can focus more on learning specialized features for more accurate face alignment. However, during training, we found that a value of 0 in the binary gate vector corrupts the update of the model parameters, since a value of 0 results in a gradient loss. To address the above issue, the proposed DSA model introduces the Improved SemHash method \cite{kaiser2018fast} to activate the binary gates $d''$, which can transmit the loss through the discrete outcome to the network.

During training process, the DSA model first turn on the $switch{\rm{ }}{\kern 1pt}{\kern 1pt}{\kern 1pt}{\kern 1pt}\varepsilon $ and a noise vector $\varepsilon $ can be combined with $d$ to get a noisy channel vector $d'$: $d'= d + \varepsilon$. The noise vector $\varepsilon $ is sampled from a Gaussian distribution with a zero mean and a unit variance and $\varepsilon \in {R^{1 \times c}}$. The noise vector $\varepsilon $ is introduced to help the model better identify feature encoding and image semantics, while making the feature encoding as close to binary as possible. Then, we use $d'$ to calculate two vectors  ${d_\alpha }$ and ${d_\beta }$ with the following equations:
\begin{equation}
	{d_\alpha } = sat\_sig(d') = \max (0,min(1,1.2\sigma (d') - 0.1))
\end{equation}
\begin{equation}
	{d_\beta } =bin\_act(d') = \left\{ {\begin{array}{*{20}{c}}
			{{\kern 1pt}{\kern 1pt}{\kern 1pt}{\kern 1pt}{\kern 1pt}1,{\rm{   }}{\kern 1pt}{\kern 1pt}{\kern 1pt}{\kern 1pt} d' > 0}\\
			{0,{\rm{     }}{\kern 1pt}{\kern 1pt}{\kern 1pt}{\kern 1pt}{\kern 1pt}{\kern 1pt}else}
	\end{array}} \right.
\end{equation} where $\sigma $ represents the sigmoid function, $sat\_sig$ and $bin\_act$ denote the saturating sigmoid function and binary activation function, respectively. ${d_\alpha }$ is a real-valued vector with values in [0,1], while d?? is a binary vector containing only 0 ones. In each iteration of the forward propagation, we randomly select half of the samples using the operation $d'' = {d_\alpha }$ and $d'' = {d_\beta }$ for the rest of the samples because we need a binary vector representing channel activation and a gradient that can be backpropagated. DSA model also follows \cite{kaiser2018fast} to set the gradient of ${d_\beta }$ w.r.t $d'$ to be the same as the gradient of ${d_\alpha }$ w.r.t $d'$ in the background propagation.

During the training stage, $switch{\rm{ }}{\kern 1pt}{\kern 1pt}{\kern 1pt}{\kern 1pt}\varepsilon $ should be turned on. In each iteration, $switch{\rm{ }}{\kern 1pt}{\kern 1pt}{\kern 1pt}{\kern 1pt}\alpha $ and $switch{\rm{ }}{\kern 1pt}{\kern 1pt}{\kern 1pt}{\kern 1pt}\beta $ will be choosed to turn on with a probability of 50\%. During the testing stage, firstly, DSA model does not need the noise $\varepsilon$. Second,  DSA model always uses the binary vector ${g_\beta }$ for predicting during the evaluation process and inference process. Therefore, the $switch{\rm{ }}{\kern 1pt}{\kern 1pt}{\kern 1pt}{\kern 1pt}\varepsilon $ and $switch{\rm{ }}{\kern 1pt}{\kern 1pt}{\kern 1pt}{\kern 1pt}\alpha $ should always be off, and the  $switch{\rm{ }}{\kern 1pt}{\kern 1pt}{\kern 1pt}{\kern 1pt}\beta $ should always be on. 

The proposed DSA model can perceive the semantic information of the face images, i.e., activate similar binary gates for the same sample subset, thus, we named it \textbf{Dynamic Semantic-Aware (DSA) model}. We believe that the dynamic learning mode in the DSA model is crucial for studying the specialized features of faces. Compared to the normal deep models \cite{Huang2021ADNetLE, Sun2024UnsupervisedMN} with shared pathways, the proposed DSA model is more adaptive to different semantic complexities. Moreover, the DSA model can effectively filter the unimportant channels and avoid the negative optimization of the channel parameters, thereby being ready for learning more discriminative and representative features.

\begin{figure*}[!t]
	\begin{center}
		\includegraphics[width=0.8\linewidth]{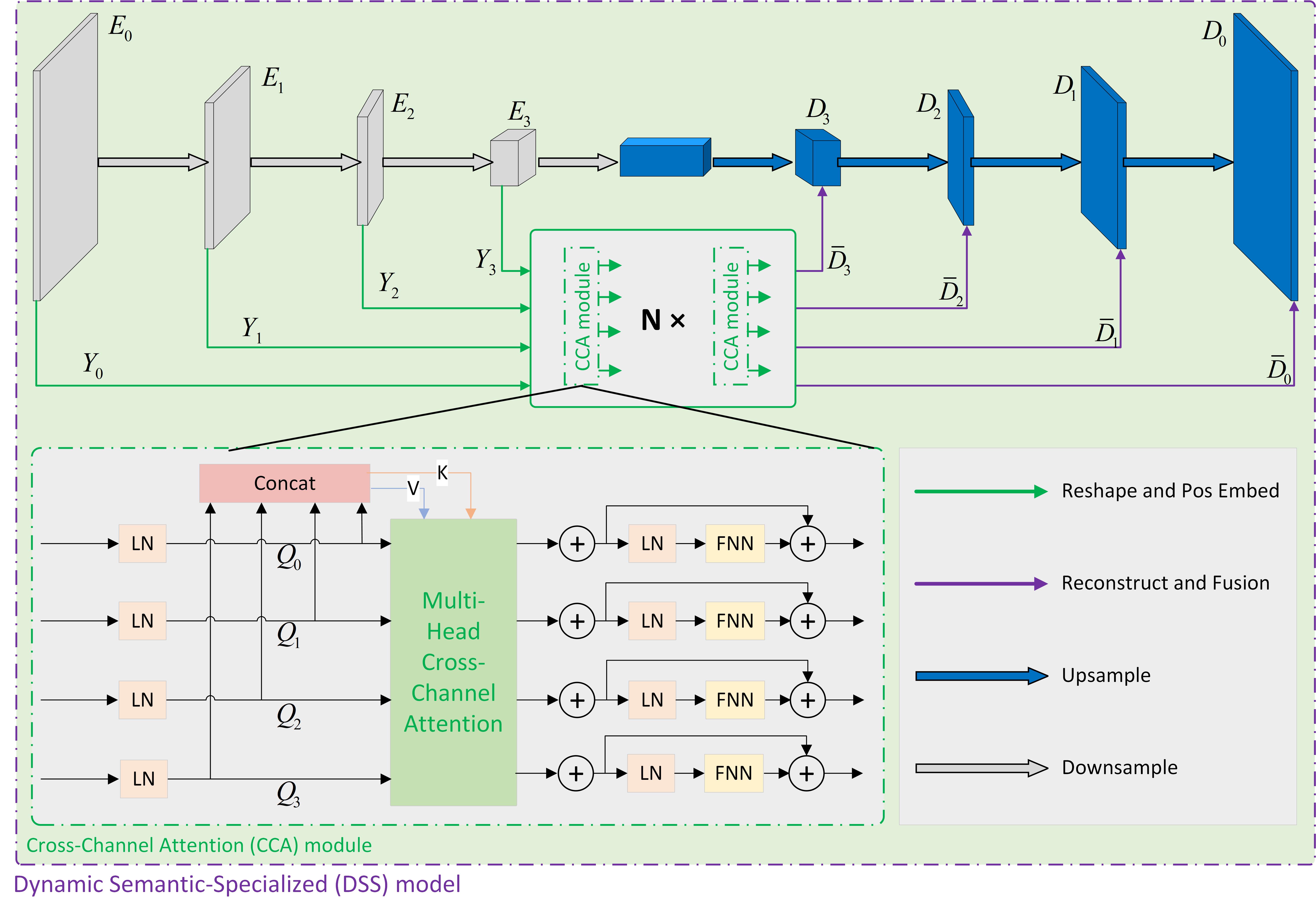}
	\end{center}
	\caption{The network architecture of the DSS model. With a well-designed Cross-Channel Attention (CCA) module, the DSS model is proposed to mine homogeneous information between them by querying features at different scales, thereby making up the semantic gap between them and eliminating semantics ambiguity. Moreover, the DSS model can also work in a dynamic feature manner to save computational costs and enhance feature representations.}
	\label{DSET}
\end{figure*}

\subsection{Dynamic Semantic Specialization model}
The proposed DSA model can effectively partition the training samples into subsets, then we need to learn the specialized features from each subset to achieve more robust face alignment. Popular works have shown that using different models to learn specialized features from different sample subsets could improve performance but also lead to massive computational cost problems. Inspired by the dynamic feature strategy in Dynamic Neural Networks, we aim to use one model to learn comparable specialized features from all sample subsets by dynamically adjusting its parameters, i.e., we propose a Dynamic Semantic Specialization (DSS) model to mine homogeneous information in features of different scales to learn specialized features and enhance feature representations.

Since the hourglass network can capture multi-scale image features, it enhances the model's feature extraction capability. This has been applied in many computer vision tasks, such as face alignment, human pose estimation, etc. The proposed DSS model also takes the Hourglass Network unit as its backbone. However, current Hourglass Networks often ignore the semantic gap among the multi-scale encoder features and between the stage of the encoder and decoder, limiting their representation ability. For example, skip connections are usually used to combine different scales of features, which introduces new noise and deteriorates the learned feature due to the semantic ambiguity of features at different scales. Therefore, the DSS model designs a Cross-Channel Attention (CCA) module (as shown in Fig. \ref{DSET}) to mine the homogeneous information from different scales of features by querying features of different scales, thereby eliminating the semantic gap and enhancing feature representations. Therefore, the proposed DSS model is capable of learning more efficient specialized features for enhancing alignment accuracy.

The proposed DSS model consists of three components, namely, the encoder, CCA module, and decoder. The input of DSS model can be the output of the DSA model (i.e., $X'$, and $X' \in {R^{W \times H \times C}}$) or the output of the last DSS model. 

\textbf{Encoder.} The encoder is able to learn multi-scale features, and the whole process can be formulated as follows:
\begin{equation}
	\left\{ {{Y_t}} \right\}_{t = 0}^3 = encoder\left( {X'} \right)
\end{equation}where $Y_t$ denote the learned features corresponding to the $t$-th scale and ${Y_t} \in {R^{\frac{W}{{{2^t}}} \times \frac{H}{{{2^t}}} \times C}}$. Then, $\left\{ {{Y_t}} \right\}_{t = 0}^3$ will be input into the Cross-Channel Attention (CCA) module for mining homogenious information among features at different scales.

\textbf{Cross-Channel Attention (CCA) module.} Since the proposed DSA model can effectively partition all training samples by estimating the semantic correlation between channels of sample feature maps, we propose a novel Cross-Channel Attention (CCA) module to mine the homogeneous information among features at different scales from the perspective of feature channels. With the proposed CCA module, feature from different scales could query and update each other from the perspective of feature channels, thereby making up the semantic gap and eliminating the semantic ambiguity between them. Specifically, the CCA module uses convolution operations with different convolution kernels to obtain the feature maps with the same dimension as $Y_3$. Then, the flatten, positional embedding and dropout operations are used to obtain the embedding sequence $Z_t$, where ${Z_t} \in {R^{\frac{{WH}}{{{{8}^2}}} \times C}}$. After that, we connect all $\left\{ {{Z_t}} \right\}_{t = 0}^3$ in channel dimension to obtain $Z_{all}$ and ${Z_{all}} \in {R^{\frac{{WH}}{{{{8}^2}}} \times 4C}}$. After that, different layer normalization (LN) operations and embedding matrixes are applied to $Z_t$ and $Z_{all}$, and the whole process can be formulated as follows:
\begin{align}
	& Z_{all} = Concat(\left\{ {{Z_t}} \right\}_{t = 0}^3) \\
	& Q_t = LN_t(Z_t)W^t_Q \\
	& K = LN(Z_{all})W_K\\
	& V =  LN(Z_{all})W_V
\end{align}where $W^t_Q \in {R^{C \times s}}$, $W_K \in {R^{4C \times s}}$ and  $W_V \in {R^{4C \times s}}$ are embedding matrixes corresponding to Query (Q), Key (K) and Values (V), respectively. $s$ is the sequence length (patch numbers). Then, a multi-head CCA model is employed to mine the homogeneous features from the perspective of feature channels with the querying attention, which can be expressed as follows:
\begin{equation}
	{{{{\bar Q}_t} = \left[ {\sum\limits_{h = 1}^H {\sigma \left( {\frac{{Q_{h,t}^{\rm{T}}{K_h}}}{{2\sqrt C }}} \right)} V_h^{\rm{T}}} \right]} \mathord{\left/
			{\vphantom {{{{\bar Q}_t} = \left[ {\sum\limits_{h = 1}^H {\sigma \left( {\frac{{Q_{h,t}^{\rm{T}}{K_h}}}{{2\sqrt C }}} \right)} V_h^{\rm{T}}} \right]} H}} \right.
			\kern-\nulldelimiterspace} H}
\end{equation}where $H$ denote the number of heads and ${{\bar Q}_t}$ means the updated feature sequence. After that, we proceed to update the obtained feature sequence using the Multi-Layer Perceptron (MLP) and residual structure to enhance the sequence, which can be expressed as follows:

\begin{equation}
	{\bar Z_{{t}}} = Z_t + MLP( LN({Z_t} + {{\bar Q_t}}) )
\end{equation}

CCA module repeats the multi-head cross-channel attention for several times. The output of the last attention ${\bar Z'_{{t}}}$ will be used to reconstruct feature maps. To be specific, we first transform the embedding squeenze ${\bar Z'_{{t}}}$ into feature maps, and reconstruct the multi-scale feature maps with reconstructtion operations. The reconstruction operation includes the upsample, convolution, batch normalization and ReLU operations. Then, the reconstructed feature maps will be added to the $Y_t$, the whole process can be formulated as follows:
\begin{equation}
	{\bar D_t} = Y_t + \sigma_1 \left( {BN\left( {conv\left( {U{P_{ \times {2^{t - 1}}}}\left( {Trans\left( {{{\bar Z}_t}} \right)} \right)} \right)} \right)} \right)
\end{equation}where $\sigma_1$ denotes the $ReLU$ activation, ${{UP}_{ \times {2^{t - 1}}}}$ means the Upsample operation, BN denotes the batch normalization and $Trans()$ means the operation that transform the the embedding sequence into feature maps. $\left\{ {{\bar D_t}} \right\}_{t = 0}^3$ contain the homogenious information among feature of different scales, which helps to eliminate the semantic gap and redundancy among feature of different scales for learn more effective specialized representations. 

\textbf{Decoder.} The reconstructed feature maps ${\bar D_t}$ will be added to each block in decoder (i.e., $D_t$ and $t=3, 2, 1, 0$ as shown in Fig. \ref{DSET}) to generate new feature maps. 

Since the DSS model aims to eliminate the semantic gap and redundancy among features of different scales from the perspective of feature channels, this further helps the DSA model to partition sample subsets from the perspective of feature channels, thereby learning more effective specialized features and improving the alignment accuracy.

\subsection{Loss Function}
{\color{red}}Inspired by \cite{Wu2018LookAB}, we introduce boundary heatmap prediction into our network as a subtask. In particular, DSAT exploits supplementary channels to process boundary information and shares a common prediction model between the boundary heatmap and the landmark heatmap. We believe that this approach will effectively capture the global features of the face, thereby strengthening the constraints of the landmarks and enhancing the quality of the landmark heatmap. Specifically, we use $\hat X$ to denote the output of the DSS model and $\hat X \in {R^{W \times H \times C}}$. Then, serval convolution, batchnorm operations and a convTranspose2d operation are used to generate the boundary and landmark heatmaps. For the supervised learning of our model, we implement the L2 loss, which can be formulated as follows: 
\begin{equation}
	Loss={{L}_{2}}(H,{{H}^{*}})+{{L}_{2}}(B,{{B}^{*}})
\end{equation}where ${H}$ and ${{{H}^{*}}}$ represent the predicted landmark heatmap and the ground truth landmark heatmap, respectively.  ${B}$ and ${{{B}^{*}}}$ represent the predicted boundary heatmap and the ground truth boundary heatmap, respectively.  

\subsection{Dynamic Semantic-Aggregation Transformer (DSAT)}
The proposed Dynamic Semantic-Aware (DSA) model can effectively partition all samples into sample subsets by estimating the semantic correlation between challenges of features and activating specific pathways for each sample subset. Then, the Dynamic Semantic-Speficialized model is designed to mine the homogenized information from features at different scales and aggregate them without semantic gaps with dynamic feature way, thereby obtaining more specialized features and saving computational cost. Finally, by integrating the DSA(Dynamic Semantic-Aware) model and DSS(Dynamic Semantic Specialization)  model into our proposed Dynamic Semantic-Aggregation Transformer (DSAT), we can learn more domain-specific features to achieve a more robust and accurate face alignment. The detailed network structure of the proposed DSAT is illustrated in Fig. \ref{overview}.

\section{Experiments}
In this section, we first introduce our experiment settings including datasets, evaluation metrics and implementation details. Then we conduct a comparative analysis against state-of-the-art methods on various benchmark datasets. Finally, we perform the self-evaluations and ablation studies for our proposed model.
\subsection{experiment setting}
\textbf{Dataset.} 
In this paper, we will conduct the experiment on the following several datasets including  ALFW \cite{Kstinger2011AnnotatedFL}, WFLW \cite{Wu2018LookAB}, 300W \cite{Sagonas2013300FI} and COFW \cite{BurgosArtizzu2013RobustFL}. 

\textbf{ALFW} dataset is a conventional dataset with diverse poses and angles of faces. Each face is annotated with 19 landmarks, and the dataset is divided into 20,000 images for training and 4,386 images for testing.

\textbf{WFLW} dataset includes 10000 images with 98 manually annotated landmarks, 7500 for training and 2500 for testing. This dataset is divided into several subclass datasets, each containing unique labels, such as heavy occlusion, make-up and illumination. We can use WFLW to evaluate our model comprehensively.

\textbf{300W} dataset is a widely used dataset in face alignment. It contains 4 subsets including AFW, HELEN, IBUG and LFPW datasets. The training phase encompasses a total of 3148 images and encompasses the complete collection of AFW, as well as the training subset derived from HELEN and LFPW. For the purpose of testing, the Fullset comprises 689 images and is subdivided into multiple subsets, namely the Common subset, which encompasses the testing subset from HELEN and LFPW, and the Challenge subset, also known as IBUG.

\textbf{COFW} dataset is a very difficult dataset because it mainly consists of images with severe occlusions and large poses. The dataset consists of 1852 images with 29 landmarks, divided into 1345 images for the training set and 507 images for the test set.

\textbf{Evaluation Metrics.} 
Following \cite{Xia2022SparseLP}, we adopt Normalized Mean Error (NME) and Failure Rate (FR) to evaluate our model, where $FR$ means the percentage of test samples whose NME are larger than a certain threshold. The NME is defined as the average Euclidean distance between the predicted facial landmarks and their corresponding ground-truth annotations, which can be defined as follows:

\begin{equation}
	NME = \frac{1}{N}\sum\limits_{{\rm{i}} = 1}^N {(\frac{{{{\left\| {{p_i} - p_{i}^*} \right\|}_2}}}{d})}  \times 100\% 
\end{equation}where $N$ is the number of landmarks in a face shape, $d$ denotes the normalization term which can be the distance between outer eye corners (inter-ocular) or the distance between pupil centers (inter-pupils). ${p_i}$ and $p_{i}^*$ represent the predicted and ground-truth landmark coordinates, respectively. 

\textbf{Implementation Details.} 
In this paper, the original input images undergo a process of cropping and resizing to attain a standardized input dimension of 256 $\times$ 256, according to the provided bounding boxes. The size of output heatmaps is 128 $\times$ 128. In order to enrich the datasets, data augmentation techniques are employed, including random horizontal flipping, grayscale transformations, occlusion effects, and rotation. The angle of rotation and bounding box scale are randomly sampled from a Gaussian distribution. We train the proposed model with Adam optimizer, and the initial learning rate is set to $2.5 \times {10^{{\rm{ - 4}}}}$. The model is trained over a period of 100000 iterations, with the learning rate being halved every 40000 iterations. 4 DSA models and 4 DSS models are stacked to construct DSAN. Each DSS model contains 3 CCA modules with 4 heads and the number of network stages is 4. The DSAN is trained with Pytorch on a single NVIDIA RTX 4090 GPU.

\begin{table}
	\caption{$\rm NME_{io}$ and $\rm NME_{ip}$ comparisions on 300W dataset. (\% omitted)}
	\footnotesize
	\begin{center}
		\begin{tabular}{p{4cm}|p{2cm}p{2cm}p{1.8cm}}
			\hline
			Method  & 
			\begin{tabular}[c]{@{}c@{}}Common\\ Subset\end{tabular} & \begin{tabular}[c]{@{}c@{}}Challenging\\ Subset\end{tabular} & Fullset \\ \hline   
			$\rm NME_{io}$ comparisions
			\\ \hline                                                  
			LAB{$\rm _{CVPR18}$}\cite{Wu2018LookAB}            & 2.98    & 5.19      & 3.49    \\
			AWing{$\rm _{ICCV19}$}\cite{Wang2019AdaptiveWL}     & 2.72  & 4.52   & 3.07 \\
			ADNet{$\rm _{ICCV21}$}\cite{Huang2021ADNetLE}       & 2.53  & 4.85    & 2.93  \\
			SLPT{$\rm _{CVPR22}$}\cite{Xia2022SparseLP}       & 2.75  & 4.90    & 3.17  \\
			DTLD{$\rm _{CVPR22}$}\cite{Li2022TowardsAF}       & 2.60  & 4.48    & 2.96  \\
			LDDMM-Face{$\rm _{PR24}$}\cite{Yang2024LDDMMFaceLD}			& 3.07  & 5.40    & 3.53  \\
			SRN+HG{$\rm _{PR22}$}\cite{ZHU2022108325}	 &3.08 &5.86 &3.62 \\
			STAR{$\rm _{CVPR23}$}\cite{Zhou2023STARLR}       & 2.52  & 4.32    & 2.87  \\
			\hline
			SHN  & 3.11  & 6.23    & 3.72  \\ 
			\textbf{SHN+DSA model }   &2.69 &4.34 &3.02\\ 
			\textbf{SHN+DSS model }  &2.70 &4.29 &3.01\\ 
			\textbf{DSAT} &\textbf{2.41} &\textbf{4.25} &\textbf{2.86}\\ 
			\hline
			$\rm NME_{ip}$ comparisions
			\\ \hline
			STKI{$\rm _{ACM MM20}$}\cite{Zhu2020SpatialTemporalKI}	&3.36	&7.39&	4.16\\
			ADNet{$\rm _{ICCV21}$}\cite{Huang2021ADNetLE}       & 3.51  & 6.47    & 4.08  \\
			DeiT-T{$\rm _{CVPR24}$}\cite{yin2024sce}       & -  & -    & 4.22  \\
			DeiT-S{$\rm _{CVPR24}$}\cite{yin2024sce}       & -  & -    & 3.94  \\
			\hline
			\textbf{DSAT} &\textbf{3.35} &\textbf{6.27} &\textbf{4.01} \\  \hline
		\end{tabular}
	\end{center}
	\label{tab300w}
	\vspace{-1em}
\end{table}

\subsection{Comparison with state-of-the-art methods}
In this part, we compare our model with the state-of-the-art methods on three different conditions to prove the robustness of the proposed DSAT. To ensure impartial assessments, we evaluate our model based on the standardized settings introduced in this section.

\begin{table*}
	\caption{$\rm NME_{io}$ comparisons on WFLW dataset. (\% omitted).}
	\begin{center}
		\footnotesize
		\begin{tabular}{p{3.5cm}|p{1.2cm}p{1.2cm}p{1.2cm}p{1.4cm}p{1.4cm}p{1.2cm}p{1.1cm}}
			\hline
			Method  &
			\begin{tabular}[l]{@{}l@{}}Testset\\\end{tabular} & \begin{tabular}[l]{@{}l@{}}Pose\\ Subset\end{tabular} &
			\begin{tabular}[c]{@{}l@{}}Expression\\ Subset\end{tabular} &
			\begin{tabular}[c]{@{}l@{}}Illumination\\ Subset\end{tabular} &
			\begin{tabular}[c]{@{}l@{}}Make-Up\\ Subset\end{tabular} &
			\begin{tabular}[c]{@{}l@{}}Occlusion\\ Subset\end{tabular} &
			\begin{tabular}[c]{@{}l@{}}Blur\\ Subset\end{tabular}  \\ \hline
			LAB{$\rm _{CVPR18}$}\cite{Wu2018LookAB} &5.27 &10.24 &5.51 &5.23 &5.15 &6.79 &6.32 \\
			MHHN{$\rm _{TIP21}$}\cite{Wan2020RobustFA} &4.77 &9.31 &4.79 &4.72 &4.59 &6.17 &5.82 \\
			LDDMM-Face{$\rm _{PR24}$}\cite{Yang2024LDDMMFaceLD} &4.63 &8.21 &5.00 &4.53 &4.31 &5.37 &5.22\\
			AWing{$\rm _{ICCV19}$}\cite{Wang2019AdaptiveWL} &4.36 &7.38 &4.58 &4.32 &4.27 &5.19 &4.96 \\
			ADNet{$\rm _{ICCV21}$}\cite{Huang2021ADNetLE} & 4.14  &6.96 & 4.38 & 4.09 & 4.05 & 5.06 &4.79 \\
			SLPT{$\rm _{CVPR22}$}\cite{Xia2022SparseLP}  & 4.14  &6.96 & 4.45 & 4.05 & 4.00 & 5.06 &4.79  \\
			\hline
			SHN &5.78 &9.47 &6.39 &5.83 &5.91 &7.07 &7.21 \\
			\textbf{DSAT} &\textbf{4.12} &\textbf{6.87} &\textbf{4.27} &\textbf{4.02} &\textbf{3.83} &\textbf{4.81} &\textbf{4.68}   \\ 
			\hline
		\end{tabular}
	\end{center}
	\label{tabwflw}
\end{table*}
\textbf{Evaluation under Normal Circumstances.} 
Under normal circumstances, the testing set should contain less variation in facial poses and occlusions, therefore, we conduct experiments on the common subset of 300W and the fullset of 300W to demonstrate the effectiveness of our proposed DSAT in normal circumstances. Table \ref{tab300w} shows the experiment results between our DSAT and several state-of-the-art face alignment models. Although these two datasets are not very challenging, our DSAT can still achieve 2.41 ${NME}_{io}$ in the common subset, and 2.86 ${NME}_{io}$ in the fullset, which outperforms state-of-the-art face alignment models. These results suggest that our DSAT can enhance detection accuracy in normal circumstances, primarily due to 1) the DSA model can dynamically activate specific pathways by estimating the semantic correlations of feature channels, helping to enhance the representation ability, 2) with the well-designed CCA module, the proposed DSS model is able to mine homogeneous information from features of different scales by querying them, alleviating the semantic gap and redundancy, and 3) by integrating the DSA model and DSS model into a novel DSAT in both dynamic architecture and dynamic parameter manners, more effective specialized features can be learned to achieve precise face alignment.

\textbf{Evaluation of Robustness against Occlusion.}
Variations in occlusion and illumination can seriously reduce the accuracy of face alignment methods. Due to the occlusion of images, it is difficult for the model to extract effective facial features, and the learned features are often accompanied by noise information. It is difficult for general models to handle such samples. In this paper, we evaluate the robustness of the proposed DSAT against occlusion on COFW dataset, 300W challenging subset and WFLW dataset, which mainly consists of challenging samples with occlusions and profile faces.

As illustrated in Table \ref{tab300w}, we compare our approach against the state-of-the-art methods on 300W challenging subset. Compared to the other methods, the proposed DSAT achieves impressive result, i.e., 4.25 in ${NME}_{io}$, which beats other state-of-the-art methods.

We also evaluate our DSAT on the WFLW dataset, which contains challenging subsets such as the make-up Subset and Occlusion Subset. As shown in Table \ref{tabwflw}, our DSAT improves by 2$\%$ and 4$\%$ in the make-up Subset and Occlusion Subset respectively compared to the best results SLPT \cite{Xia2022SparseLP}. These experimental results indicate that our proposed DSAT is more robust to faces with complicated occlusions.

On the COFW dataset, the failure rate is defined as the percentage of test samples with prediction errors exceeding 10\%. The comparison results of NME and FR  between our DSAT and state-of-the-art methods are reported in Table \ref{tabcofw}. On the COFW dataset, our proposed DSAT reaches an excellent result with 2.57 ${NME}_{io}$ and 0.00 FR, which outperforms the state-of-the-art methods. The result of ${NME}_{ip}$ is 4.74, which is a huge improvement.

\begin{table}
	\caption{$\rm NME_{ip}$ and $\rm FR_{ip}^{10}$ comparisons on COFW dataset. (- not counted, \% omitted)}
	\begin{center}
		\footnotesize
		\begin{tabular}{p{4.6cm}|p{1.2cm}p{1.2cm}}
			\hline
			Method & $\rm NME_{ip}$  & $\rm FR_{ip}^{10}$ \\
			\hline
			LAB{$\rm _{CVPR18}$}\cite{Wu2018LookAB}&	5.58&	2.76\\
			AWing{$\rm _{ICCV19}$}\cite{Wang2019AdaptiveWL}&	4.94&	0.99\\
			MHHN{$\rm _{TIP21}$}\cite{Wan2020RobustFA} &4.95 &1.78 \\
			SRN {$\rm _{PR22}$}\cite{ZHU2022108325}	 &4.85 &- \\
			LDDMM-Face{$\rm _{PR24}$}\cite{Yang2024LDDMMFaceLD} &4.54 &1.18 \\
			\hline
			SHN &	6.21&	5.52\\
			\textbf{DSAT} &\textbf{4.74}  &\textbf{0.00} \\ 
			\hline
			&$\rm NME_{io}$  & $\rm FR_{ip}^{10}$ \\
			\hline
			LAB{$\rm _{CVPR18}$}\cite{Wu2018LookAB}&3.92 &	0.39\\
			ADNet{$\rm _{ICCV21}$}\cite{Huang2021ADNetLE} & 2.68  & 0.59 \\
			DTLD{$\rm _{CVPR22}$}\cite{Li2022TowardsAF}  & 3.02  & -  \\
			SLPT{$\rm _{CVPR22}$}\cite{Xia2022SparseLP}  & 3.32  & \textbf{0.00}  \\
			\hline
			\textbf{DSAT} &\textbf{2.57}  &\textbf{0.00} \\ 
			\hline
		\end{tabular}
	\end{center}
	\label{tabcofw}
\end{table}

Hence, from the above experiments, we can find that DSAT can be more robust against occlusion. This is mainly because 1) The DSA model can perceive the semantics of samples, and effectively activate the specific pathways to remove the noise information from occlusions, 2) our DSS model can alleviate semantic gaps and redundancy among features of different scales for learning their homogeneous information, and 3) by integrating the above two dynamic models, our proposed DSAT can learn more effective specialized feature for detecting more accurate landmarks.

\textbf{Evaluation of Robustness against Large Poses.}
Faces with various poses are another great challenge for face alignment methods. For large pose images of human faces, the location of landmarks and their relationships vary greatly, thereby increasing the difficulty of face alignment. In order to assess the performance of our proposed DSAT method for various poses, we choose to conduct experiments on AFLW dataset, 300W challenging subset and WFLW dataset.

\begin{table}
	\caption{$\rm NME$ and $\rm AUC$ comparisions on AFLW dataset. (- not counted, \% omitted)}
	\begin{center}
		\footnotesize
		\begin{tabular}{p{4.6cm}|p{1.2cm}p{1.2cm}}
			\hline
			\multirow{2}{*}{Method} & {$\rm NME_{diag}$} \\
			\cline{2-3}
			& Full & Frontal \\
			\cline{2-3}	
			\hline
			LLL{$\rm _{ICCV19}$}\cite{Robinson2019LaplaceLL} & 1.97 & -   \\
			LAB{$\rm _{CVPR18}$}\cite{Wu2018LookAB} & 1.85 & 1.62  \\
			MHHN{$\rm _{TIP21}$}\cite{Wan2020RobustFA} & 1.38 &1.19   \\ 
			DTLD{$\rm _{CVPR22}$}\cite{Li2022TowardsAF}  & 1.37  & -  \\
			\hline
			SHN &	2.46&	1.92\\
			\textbf{DSAT} &\textbf{1.35} &\textbf{1.17}   \\ 
			\hline
		\end{tabular}
	\end{center}
	\label{tabaflw}
\end{table}

The comparison results between our DSAT and the state-of-the-art methods on the AFLW dataset are presented in Table \ref{tabaflw}, and our DSAT achieves 1.35 ${NME}_{diag}$ in the Full subset and 1.17 ${NME}_{diag}$ in the Frontal subset, which outperforms the other methods. Table \ref{tab300w} and Table \ref{tabwflw} show the results on the 300W challenging subset and  WFLW dataset respectively. From the above experimental results, we can conclude that our DSAT is more robust to faces with large poses because our DSAT can adapt to faces with various samples and learn specialized features from them, thereby improving the alignment accuracy.

\subsection{Self Evaluations}
\textbf{Number and position of DSA model.} The proposed DSAN uses multiple DSA models, and different positions and numbers of DSA models may affect the final alignment accuracy. To figure this out, we conduct comprehensive experiments on the 300W dataset by using one DSA model before the first DSS model (i.e., first), one DSA model before the first DSS model and one DSA model before the second DSS model (i.e., first and second),  and four DSA models separately before four DSS models (i.e., before each DSS model), respectively. From Table \ref{tabdsa}, we can find that using one DSA model before the first and the second DSA model (i.e., first and second) obtains the best results. This is exactly in line with our assumption that the sample should be gradually partitioned for learning specialized features. In addition, as the output of our DSA model is a binary vector, and too many zeros in the vector may destroy the backpropagation of the gradient and reduce its performance.

\begin{table}
	\caption{ $\rm NME$ comparisions of different numbers and positions of the DSA model on the 300W dataset}
	\begin{center}
		\footnotesize
		\begin{tabular}{p{5.6cm}p{1.0cm}}
			\hline
			Numbers and location of DSA model & $\rm NME_{ip}$   \\
			\hline
			\textbf{1 first }  &6.27 \\ 
			\textbf{2 first and second }   &6.24\\ 
			\textbf{4 before each}   &6.50\\ 
			\hline
			numbers of attention heads & $\rm NME_{ip}$   \\
			\hline
			\textbf{2  }  &6.41 \\ 
			\textbf{4  }  &6.24\\ 
			\textbf{8  }  &6.38 \\ 
			\hline
			Numbers of CCA modules & $\rm NME_{ip}$   \\
			\hline
			\textbf{1} &6.38 \\
			\textbf{2} &6.28  \\ 
			\textbf{3} &6.24 \\ 
			\textbf{4} &6.41  \\ 
			\hline
		\end{tabular}
	\end{center}
	\label{tabdsa}
\end{table}

\textbf{Numbers of the heads.}
This section aims to explore the impact of head numbers of the DSS model on the prediction effect of the model. The corresponding result is shown in Table \ref{tabdsa}, from which we can find that a moderate number of heads is most effective for face alignment. This may be because too many heads will increase the redundancy of information, thereby reducing the representation ability of features.

\textbf{Numbers of CCA modules.}
The depth of CCA modules affects the aggregation of multi-scale features and the accuracy of the face alignment model. In order to explore the number of CCA modules with the best aggregation effect, we conduct experiments on the 300W dataset by setting the number of blocks as 1, 2, 3, and 4, respectively. As shown in Table \ref{tabdsa}, 3 CCA modules achieve the best performance. 

\begin{figure}[!t]
	\begin{center}
		\includegraphics[width=0.65\linewidth]{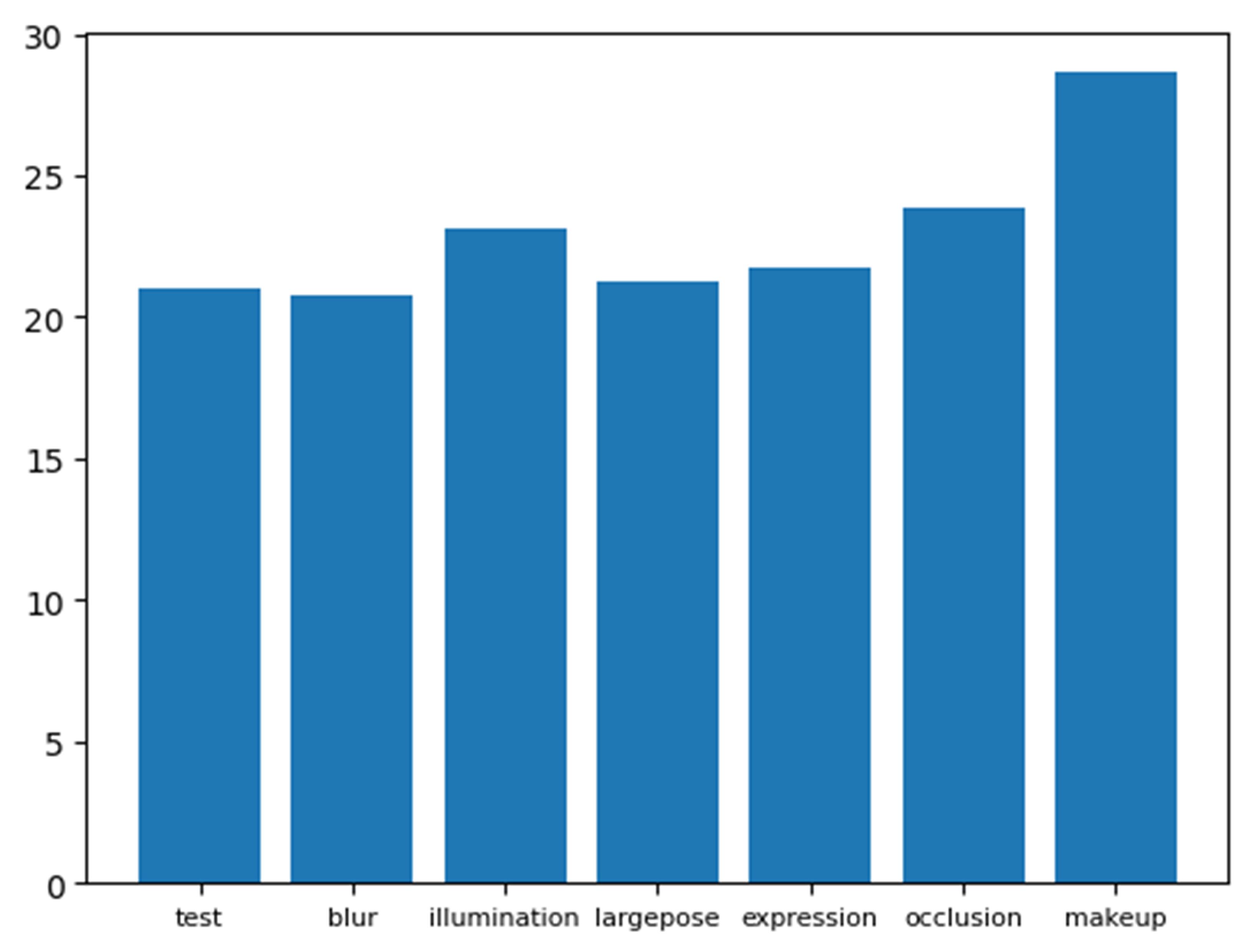}
	\end{center}
	\caption{The image shows the average number of DSA model channels activated on WLFW datasets. For different types of images, the DSA model will activate different numbers of model channels.}
	\label{gatewlfw}
\end{figure}

\textbf{Result on subsets division.}
In order to verify the effectiveness of the DSA model in subset partitioning, we conduct corresponding experiments on the WLFW dataset. We first calculate the average numbers of activated channels on WFLW dataset, and the corresponding experiment results are shown in Fig. \ref{gatewlfw}. We find that different types of samples (i.e., different WFLW testing subsets) activate different numbers of channels. For example, the number of channels activated by the illumination subset and the make-up subset is higher than other sample subsets. This may be because the illumination or the make-up hinders the learning of efficient features, driving the DSA model to activate more channels.

\begin{figure*}[!t]
	\begin{center}
		\includegraphics[width=0.65\linewidth]{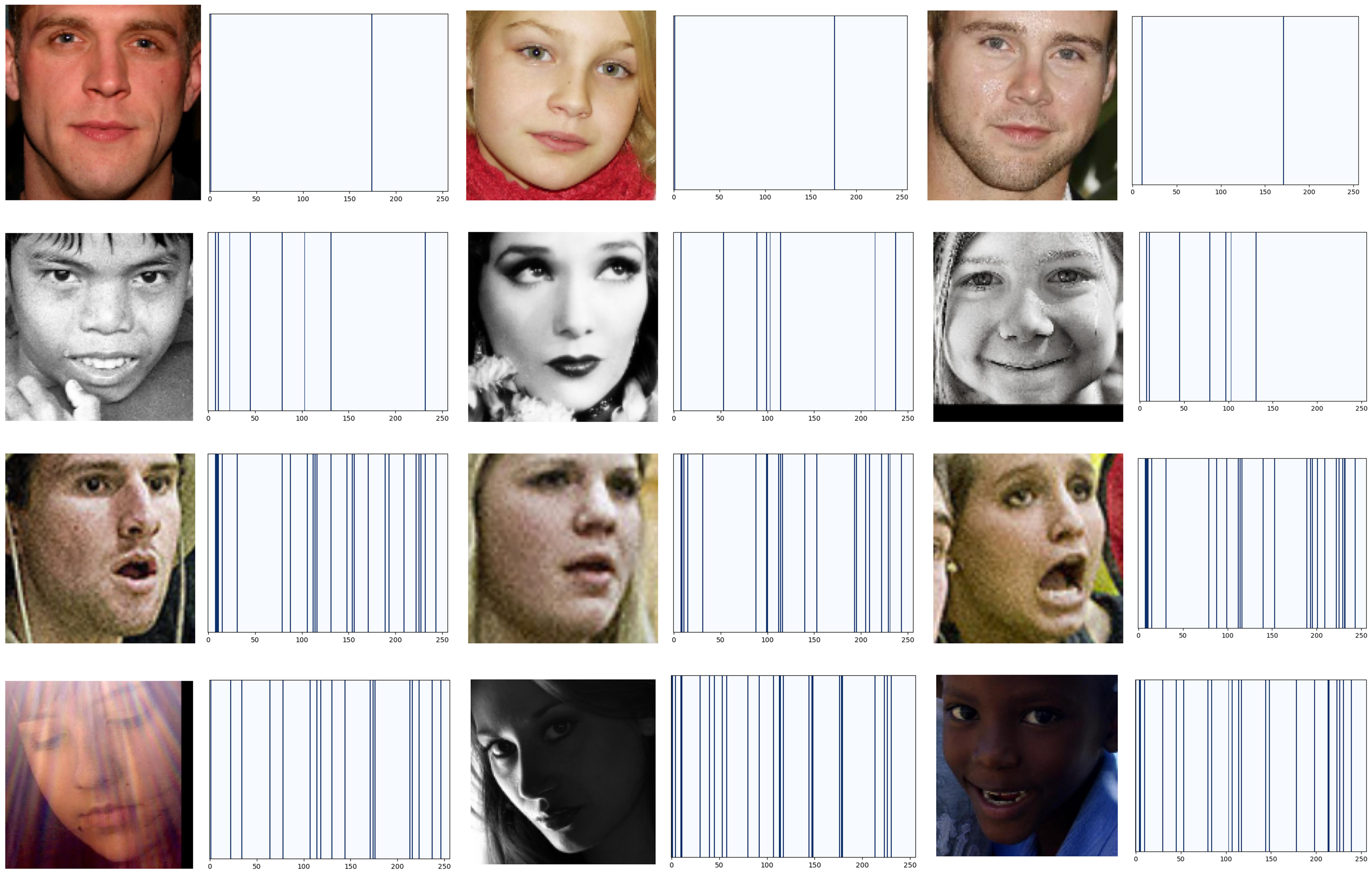}
	\end{center}
	\caption{Channel activation in different cases on 300W challenging subset. The DSA model activates different channels according to the estimated sample difficulty. The harder the sample, the more channels are activated.}
	\label{gate}
\end{figure*}

\begin{figure}[!t]
	\begin{center}
		\includegraphics[width=0.8\linewidth]{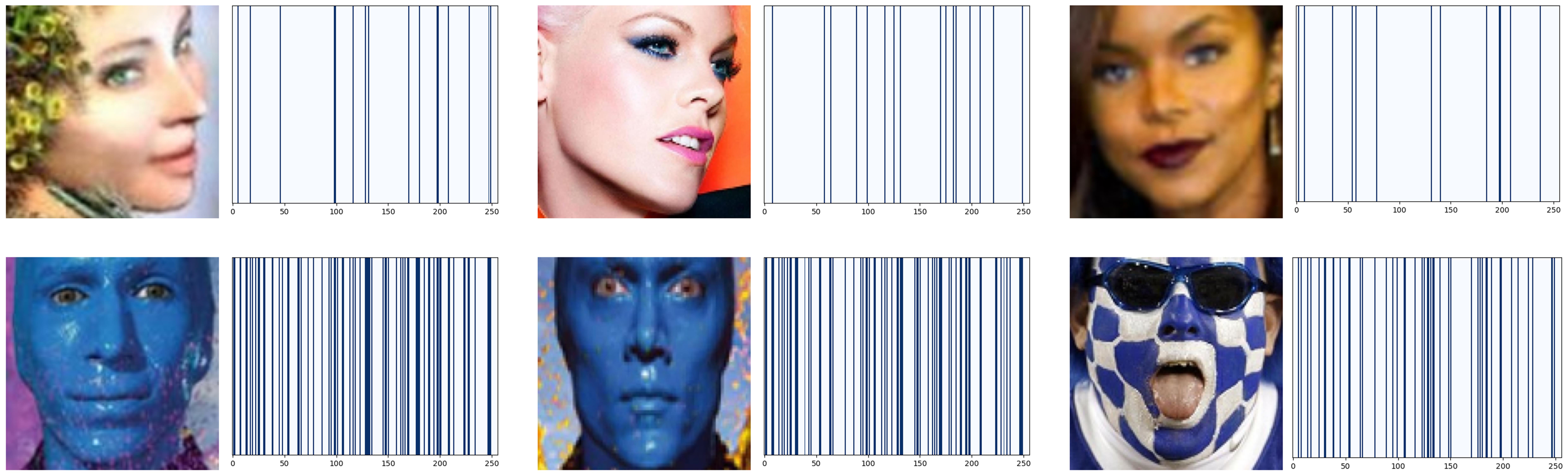}
	\end{center}
	\caption{Channel activation of different samples in WFLW make-up subset. The heavy makeup face activates more channels than the light makeup one.}
	\label{makeup}
\end{figure}

Next, we further analyze detailed channel activations. As shown in Fig. \ref{gate} and \ref{makeup}, different rows show images with similar faces and their corresponding channel activations, from which we can find that the DSA model activates similar channels for similar faces. Moreover, we can also find that the DSA model will activate different channels according to the estimated sample difficulty. For example, in Fig. \ref{gate}, the illumination faces or large pose faces activate more channels than neutral faces. In Fig. \ref{makeup}, the heavy makeup face activates more channels than the light makeup one (see Fig. \ref{makeup}). For easy samples, fewer channels are usually activated, since the model can easily detect their landmarks without the help of additional channel information. For difficult samples, more channels need to be activated to provide more semantic or spatial information to enhance feature representation. We can also find from Fig. \ref{gate}, that grayscale images rarely activate the second half of the channel. In contrast, blurred and illuminated samples use this part of the channel information. These experimental results indicate our proposed DSA model can partition all samples into subsets and activate specific channels for them to help learn specialized features.

\subsection{Ablation Studies}
The proposed DSAT mainly includes two parts: DSA model and DSS model. In order to further verify the effectiveness of these two modules, we perform the ablation study for our proposed models on the 300W dataset.

(1) The DSA model is able to distinguish the semantic information of samples and activate specific channels for them, which is beneficial for the DSAT to learn difficult samples without losing the accuracy of general samples. To evaluate this, we conduct experiments on the baseline (SHN) on the 300W dataset. As shown in Table \ref{tab300w}, the result of SHN + DSA model outperforms SHN, especially on challenging subset, indicating that our model can detect more accurate landmarks for difficult images.

\begin{figure}[!t]
	\begin{center}
		\includegraphics[width=0.80\linewidth]{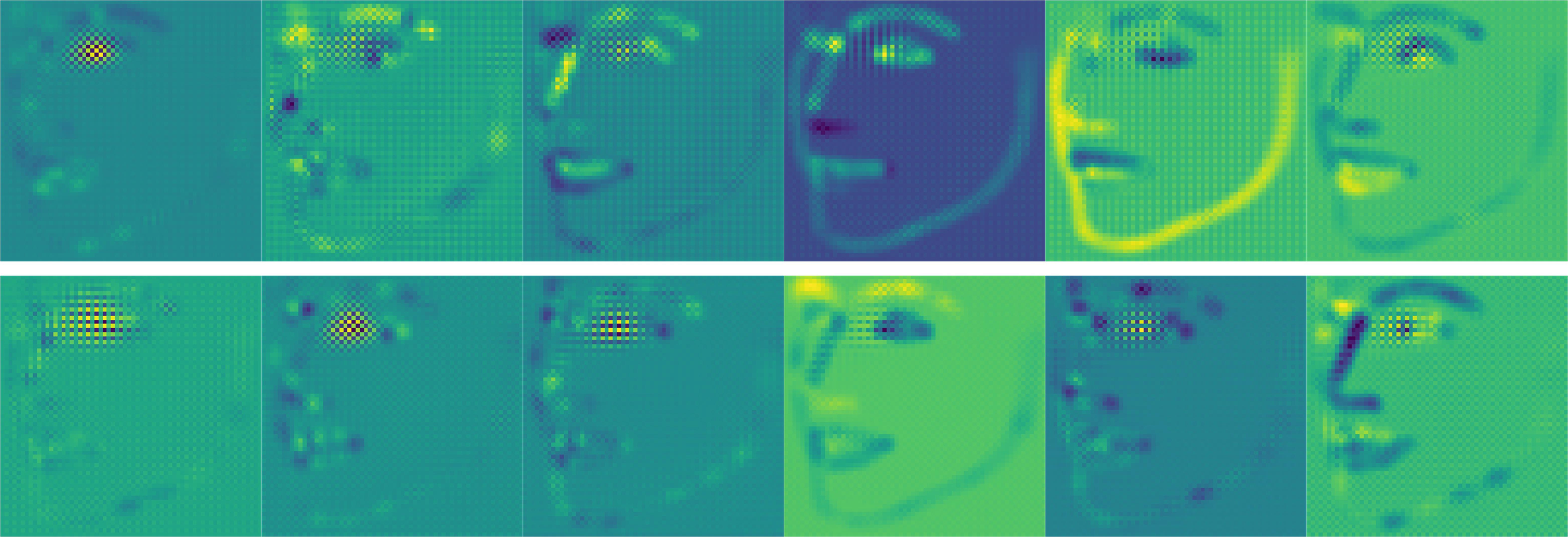}
	\end{center}
	\caption{Partial feature maps generated by DSAT with (first line) and without the DSS model (second line). We can find that by combining the DSS model, the aggregated multi-scale features alleviate the semantic gaps among them and are more discriminative and representative.}
	\label{heatmap}
\end{figure}

(2) To alleviate the semantic gap and redundancy of features at different scales for specialized feature learning, we propose the DSS model. Compared with other models using multi-scale features, our model can use the dynamic parameter strategy to explore the relationship between features and alleviate semantic confusion. We conduct the experiment by combining DSS model and SHN and the corresponding generated heatmaps are shown in Fig. \ref{heatmap}. Comparing the heatmaps generated by the two models, the heatmaps generated with the DSS model produce clearer boundary information with continuous and less obscured boundary contours. This demonstrates that our model can produce more informative heatmaps. At the same time, the former produces a smaller range of thermal values of the landmarks, which can exclude the interference information when predicting the final results. From Table \ref{tab300w}, we can find that DSS+SHN outperforms SHN, which indicates that the proposed DSS model can better aggregate multi-scale features to improve alignment accuracy.

(3) When we integrate the DSA model and DSS model into a novel DSAT framework, the alignment accuracy can be further improved, which can be seen from Table \ref{tab300w} (i.e., DSAT outperforms SHN+DSA and SHN+DSS). This experimental result indicates that the DSA model and DSS model collaborate with each other to learn more effective specialized features and improve alignment accuracy.

\subsection{ Experimental Results and Discussions}
From the above experiments and comparisons, we draw the following conclusions and findings:

(1) Compared with other heatmap regression methods\cite{ Huang2021ADNetLE, Ma2022RobustFA, wan2023precise}, our model outperforms them. The main reasons are as follows: 1) the DSA model can effectively divide the sample subset so that our proposed DSAT works well on general faces as well as hard ones, 2) the DSS model can make better use of multi-scale features by adopting dynamic parameter strategy to alleviate their semantic gaps for enhancing feature representations, and 3) by collaborating the DSA model and DSS model, more effective specialized features can be learned to improve alignment accuracy.


(2)	Many existing face alignment methods \cite{Wang2019AdaptiveWL, Huang2021ADNetLE, Gao2021AFL} use multi-scale features to improve model performance, but they ignore the influence of semantic ambiguity of features at different scales. Our proposed DSAT outperforms them, which indicates the proposed DSS model can learn more effective multi-scale features by alleviating their semantic gaps and redundancy for enhancing feature representations.

(3)	Compared to the current face alignment models \cite{Wu2018LookAB, Huang2021ADNetLE}, our proposed DSAT adds some parameters due to the use of CCA module. Our DSAT can optimize the whole model and outperform state-of-the-art models as the DSAT only needs to pay attention to updating a part of the parameters each time, and these selected channels are exactly the parameters that contribute the most to the prediction of such samples. Therefore, the proposed DSAT not only solves the parameter updating problem, but also enhances its generalization ability, leading to more precise face alignment.
{\color{red}}

\section{Conclusion}
Unconstrained face alignment remains challenging due to mediocre feature learning. In this paper, we aim to address this problem by proposing a novel Dynamic Semantic Aggregation Transformer for more discriminative and representative feature learning. Our method improves the accuracy of landmark detection compared to existing techniques, as 1) DSAT effectively partitions all samples into subsets and activates unique computational paths for each subset of samples according to their abstract semantics, which avoids the mediocrity in learning that can occur with models that use a predefined fixed structure to train all samples, and 2) DSAT learns specialized features from the sample subsets and eliminates semantic differences in different specialized features, which contributes to more effective feature learning. The enhanced performance of our approach demonstrates its potential for real-world applications, such as facial recognition, expression analysis, and augmented reality. The ability to accurately detect facial landmarks under various conditions can significantly improve the robustness and reliability of these applications.

Despite the promising results, our method has certain limitations, such as introducing the cross-channel attention (CCA) module into the proposed dynamic semantic specific (DSS) model also incurs some computational cost. Future research could focus on addressing these limitations by exploring multi-dataset joint training method or integrating additional cues, such as depth information, to handle challenging scenarios. 

\leftline{ {\bf Acknowledgements}} This work is supported by the National Natural Science Foundation of China (Grant No. 62002233 and 62262069), the Natural Science Foundation of Hubei Province, China (Grant No. 2024AFB992), and the Fundamental Research Funds for the Central Universities, Zhongnan University of Economics and Law (Grant No. 2722023BQ058)).



\quad


\leftline{ {\bf References}}

\bibliography{mybibfile}

\begin{thebibliography}{10}
\expandafter\ifx\csname url\endcsname\relax
  \def\url#1{\texttt{#1}}\fi
\expandafter\ifx\csname urlprefix\endcsname\relax\def\urlprefix{URL }\fi
\expandafter\ifx\csname href\endcsname\relax
  \def\href#1#2{#2} \def\path#1{#1}\fi

\bibitem{Wright2009RobustFR}
J.~Wright, A.~Y. Yang, A.~Ganesh, S.~Sastry, Y.~Ma, Robust face recognition via
  sparse representation, IEEE Transactions on Pattern Analysis and Machine
  Intelligence 31 (2009) 210--227.

\bibitem{He2024EnhancingFR}
M.~He, J.~Zhang, S.~Shan, X.~Chen, Enhancing face recognition with detachable
  self-supervised bypass networks, IEEE Transactions on Image Processing 33
  (2024) 1588--1599.

\bibitem{Yang2022HeterogeneousFR}
Z.~Yang, J.~Q. Liang, C.~Fu, M.~Luo, X.~Zhang, Heterogeneous face recognition
  via face synthesis with identity-attribute disentanglement, IEEE Transactions
  on Information Forensics and Security 17 (2022) 1344--1358.

\bibitem{liu2023cross}
T.~Liu, J.~Li, J.~Wu, L.~Zhang, S.~Zhao, J.~Chang, J.~Wan, Cross-domain facial
  expression recognition via disentangling identity representation., in: IJCAI,
  2023, pp. 1213--1221.

\bibitem{Savchenko2022ClassifyingEA}
A.~Savchenko, L.~V. Savchenko, I.~Makarov, Classifying emotions and engagement
  in online learning based on a single facial expression recognition neural
  network, IEEE Transactions on Affective Computing 13 (2022) 2132--2143.

\bibitem{Zhao2019MobileFANTD}
Y.~Zhao, Y.~Liu, C.~Shen, Y.~Gao, S.~Xiong, Mobilefan: Transferring deep hidden
  representation for face alignment, Pattern Recognit. 100 (2019) 107114.

\bibitem{wu2024diffusion}
T.~Wu, K.~Wang, C.~Tang, J.~Zhang, Diffusion-based network for unsupervised
  landmark detection, Knowledge-Based Systems 292 (2024) 111627.

\bibitem{tian2024adaptive}
Y.~Tian, D.~Su, S.~Li, Adaptive robust loss for landmark detection, Information
  Fusion 101 (2024) 102013.

\bibitem{Johnson2019SurveyOD}
J.~M. Johnson, T.~M. Khoshgoftaar, Survey on deep learning with class
  imbalance, Journal of Big Data 6 (2019) 1--54.

\bibitem{Han2021DynamicNN}
Y.~Han, G.~Huang, S.~Song, L.~Yang, H.~Wang, Y.~Wang, Dynamic neural networks:
  A survey, IEEE Transactions on Pattern Analysis and Machine Intelligence 44
  (2021) 7436--7456.

\bibitem{Xie2019OnGR}
Z.~Xie, L.~Jin, X.~Du, X.~Xiao, H.~Li, S.~Li, On generalized rmp scheme for
  redundant robot manipulators aided with dynamic neural networks and nonconvex
  bound constraints, IEEE Transactions on Industrial Informatics 15 (2019)
  5172--5181.

\bibitem{Cootes1995ActiveSM}
T.~Cootes, C.~J. Taylor, D.~H. Cooper, J.~Graham, Active shape models-their
  training and application, Comput. Vis. Image Underst. 61 (1995) 38--59.

\bibitem{Cootes1998ActiveAM}
T.~Cootes, G.~J. Edwards, C.~J. Taylor, Active appearance models, in: European
  Conference on Computer Vision, 1998.

\bibitem{ZHU2022108325}
C.~Zhu, X.~Li, J.~Li, S.~Dai, W.~Tong, Reasoning structural relation for
  occlusion-robust facial landmark localization, Pattern Recognition 122 (2022)
  108325.

\bibitem{liang2024generalizable}
J.~Liang, H.~Liu, H.~Xu, D.~Luo, Generalizable face landmarking guided by
  conditional face warping, in: Proceedings of the IEEE/CVF Conference on
  Computer Vision and Pattern Recognition, 2024, pp. 2425--2435.

\bibitem{Zhou2023STARLR}
Z.~Zhou, H.~Li, H.~Liu, N.~na~Wang, G.~Yu, R.~Ji, Star loss: Reducing semantic
  ambiguity in facial landmark detection, 2023 IEEE/CVF Conference on Computer
  Vision and Pattern Recognition (CVPR) (2023) 15475--15484.

\bibitem{wan2023precise}
J.~Wan, J.~Liu, J.~Zhou, Z.~Lai, L.~Shen, H.~Sun, P.~Xiong, W.~Min, Precise
  facial landmark detection by reference heatmap transformer, IEEE Transactions
  on Image Processing 32 (2023) 1966--1977.

\bibitem{Cristinacce2006FeatureDA}
D.~Cristinacce, T.~Cootes, Feature detection and tracking with constrained
  local models, in: British Machine Vision Conference, 2006.

\bibitem{Liu2009DiscriminativeFA}
X.~Liu, Discriminative face alignment, IEEE Transactions on Pattern Analysis
  and Machine Intelligence 31 (2009) 1941--1954.

\bibitem{Lin2021StructureCoherentDF}
C.~Lin, B.~Zhu, Q.~Wang, R.~Liao, C.~Qian, J.~Lu, J.~Zhou, Structure-coherent
  deep feature learning for robust face alignment, IEEE Transactions on Image
  Processing 30 (2021) 5313--5326.

\bibitem{Xu2020AnchorFaceAA}
Z.~Xu, B.~Li, M.~Geng, Y.~Yuan, G.~Yu, Anchorface: An anchor-based facial
  landmark detector across large poses, in: AAAI Conference on Artificial
  Intelligence, 2020.

\bibitem{Xia2022SparseLP}
J.~Xia, W.~Qu, W.-F. Huang, J.~Zhang, X.~Wang, M.~Xu, Sparse local patch
  transformer for robust face alignment and landmarks inherent relation
  learning, 2022 IEEE/CVF Conference on Computer Vision and Pattern Recognition
  (CVPR) (2022) 4042--4051.

\bibitem{Yang2024LDDMMFaceLD}
H.~Yang, J.~Lyu, P.~Cheng, R.~Tam, X.~Tang, Lddmm-face: Large deformation
  diffeomorphic metric learning for cross-annotation face alignment, Pattern
  Recognition (2024).

\bibitem{Wang2019AdaptiveWL}
X.~Wang, L.~Bo, F.~Li, Adaptive wing loss for robust face alignment via heatmap
  regression, 2019 IEEE/CVF International Conference on Computer Vision (ICCV)
  (2019) 6970--6980.

\bibitem{Ma2022RobustFA}
J.~Ma, J.~Li, B.~Du, J.~Wu, J.~Wan, Y.~Xiao, Robust face alignment by
  dual-attentional spatial-aware capsule networks, Pattern Recognit. 122 (2022)
  108297.

\bibitem{Huang2021ADNetLE}
Y.~Huang, H.~Yang, C.~Li, J.~Kim, F.~Wei, Adnet: Leveraging error-bias towards
  normal direction in face alignment, 2021 IEEE/CVF International Conference on
  Computer Vision (ICCV) (2021) 3060--3070.

\bibitem{kaiser2018fast}
L.~Kaiser, S.~Bengio, A.~Roy, A.~Vaswani, N.~Parmar, J.~Uszkoreit, N.~Shazeer,
  Fast decoding in sequence models using discrete latent variables, in:
  International Conference on Machine Learning, ICML, 2018, pp. 2390--2399.

\bibitem{Sun2024UnsupervisedMN}
H.~Sun, Z.~Luo, D.~Ren, B.~Du, L.~Chang, J.~Wan, Unsupervised multi-branch
  network with high-frequency enhancement for image dehazing, Pattern
  Recognition (2024).

\bibitem{Wu2018LookAB}
W.~Wu, C.~Qian, S.~Yang, Q.~Wang, Y.~Cai, Q.~Zhou, Look at boundary: A
  boundary-aware face alignment algorithm, 2018 IEEE/CVF Conference on Computer
  Vision and Pattern Recognition (2018) 2129--2138.

\bibitem{Kstinger2011AnnotatedFL}
M.~K{\"o}stinger, P.~Wohlhart, P.~M. Roth, H.~Bischof, Annotated facial
  landmarks in the wild: A large-scale, real-world database for facial landmark
  localization, 2011 IEEE International Conference on Computer Vision Workshops
  (ICCV Workshops) (2011) 2144--2151.

\bibitem{Sagonas2013300FI}
C.~Sagonas, G.~Tzimiropoulos, S.~Zafeiriou, M.~Pantic, 300 faces in-the-wild
  challenge: The first facial landmark localization challenge, 2013 IEEE
  International Conference on Computer Vision Workshops (2013) 397--403.

\bibitem{BurgosArtizzu2013RobustFL}
X.~P. Burgos-Artizzu, P.~Perona, P.~Doll{\'a}r, Robust face landmark estimation
  under occlusion, 2013 IEEE International Conference on Computer Vision (2013)
  1513--1520.

\bibitem{Li2022TowardsAF}
H.~Li, Z.~Guo, S.-M. Rhee, S.~J. Han, J.-J. Han, Towards accurate facial
  landmark detection via cascaded transformers, 2022 IEEE/CVF Conference on
  Computer Vision and Pattern Recognition (CVPR) (2022) 4166--4175.

\bibitem{Zhu2020SpatialTemporalKI}
C.~Zhu, X.~Li, J.~Li, G.~Ding, W.~Tong, Spatial-temporal knowledge integration:
  Robust self-supervised facial landmark tracking, Proceedings of the 28th ACM
  International Conference on Multimedia (2020).

\bibitem{yin2024sce}
K.~Yin, V.~Rao, R.~Jiang, X.~Liu, P.~Aarabi, D.~B. Lindell, Sce-mae: Selective
  correspondence enhancement with masked autoencoder for self-supervised
  landmark estimation, in: Proceedings of the IEEE/CVF Conference on Computer
  Vision and Pattern Recognition, 2024, pp. 1313--1322.

\bibitem{Wan2020RobustFA}
J.~Wan, Z.~Lai, J.~Liu, J.~Zhou, C.~Gao, Robust face alignment by multi-order
  high-precision hourglass network, IEEE Transactions on Image Processing 30
  (2020) 121--133.

\bibitem{Robinson2019LaplaceLL}
J.~P. Robinson, Y.~Li, N.~Zhang, Y.~R. Fu, S.~Tulyakov, Laplace landmark
  localization, 2019 IEEE/CVF International Conference on Computer Vision
  (ICCV) (2019) 10102--10111.

\bibitem{Gao2021AFL}
P.~Gao, K.~Lu, J.~Xue, J.~Lyu, L.~Shao, A facial landmark detection method
  based on deep knowledge transfer, IEEE Transactions on Neural Networks and
  Learning Systems 34 (2021) 1342--1353.

\end{thebibliography}

\end{document}